\documentclass{article}

\usepackage{microtype}
\usepackage{graphicx}
\usepackage{subfigure}
\usepackage{booktabs} 
\usepackage{soul}

\usepackage{hyperref}

\usepackage[accepted]{icml2024}

\usepackage{amsmath}
\usepackage{amssymb}
\usepackage{mathtools}
\usepackage{amsthm}

\usepackage[whole]{bxcjkjatype}
\usepackage{booktabs}
\usepackage{multirow}
\usepackage{float}
\usepackage{stfloats}
\usepackage{bm}
\usepackage{authblk}
\usepackage{url}

\usepackage{amsfonts}
\usepackage{mathrsfs}
\usepackage{color}
\usepackage[capitalize,noabbrev]{cleveref}

\theoremstyle{plain}
\newtheorem{theorem}{Theorem}[section]
\newtheorem{proposition}[theorem]{Proposition}
\newtheorem{lemma}[theorem]{Lemma}

\theoremstyle{definition}

\theoremstyle{remark}

\usepackage[textsize=tiny]{todonotes}

\icmltitlerunning{Flexible Variational Information Bottleneck: Achieving Diverse Compression with a Single Training}
\title{Flexible Variational Information Bottleneck:\\ Achieving Diverse Compression with a Single Training}
\author[1]{Sota Kudo}
\author[1]{Naoaki Ono}
\author[1]{Shigehiko Kanaya}
\author[1, 2]{Ming Huang}
\affil[1]{Nara Institute of Science and Technology}
\affil[2]{Nagoya City University}
\begin{document}
\DeclarePairedDelimiter{\abs}{\lvert}{\rvert} 
\DeclarePairedDelimiter{\norm}{\lVert}{\rVert} 
\DeclarePairedDelimiter{\rbra}{\lparen}{\rparen} 
\DeclarePairedDelimiter{\cbra}{\lbrace}{\rbrace} 
\DeclarePairedDelimiter{\sbra}{\lbrack}{\rbrack} 
\DeclarePairedDelimiter{\abra}{\langle}{\rangle} 
\DeclarePairedDelimiter{\floor}{\lfloor}{\rfloor} 
\DeclarePairedDelimiter{\ceil}{\lceil}{\rceil} 

\maketitle
\renewcommand{\thefootnote}{\fnsymbol{footnote}}
\footnote[0]{Correspondence to: Sota Kudo \textless kudo.sota.ko2@is.naist.jp\textgreater, Ming Huang \textless alex-mhuang@is.naist.jp\textgreater, \textless alex-mhuang@ds.nagoya-cu.ac.jp\textgreater, \textless alex\_mhuang@ieee.org\textgreater}
\renewcommand{\thefootnote}{\arabic{footnote}}

\begin{abstract}
Information Bottleneck (IB) is a widely used framework that enables the extraction of information related to a target random variable from a source random variable. In the objective function, IB controls the trade-off between data compression and predictiveness through the Lagrange multiplier $\beta$. Traditionally, to find the trade-off to be learned, IB requires a search for $\beta$ through multiple training cycles, which is computationally expensive. In this study, we introduce Flexible Variational Information Bottleneck (FVIB), an innovative framework for classification task that can obtain optimal models for all values of $\beta$ with single, computationally efficient training. We theoretically demonstrate that  across all values of reasonable $\beta$, FVIB can simultaneously maximize an approximation of the objective function for Variational Information Bottleneck (VIB), the conventional IB method. Then we empirically show that FVIB can learn the VIB objective as effectively as  VIB. Furthermore, in terms of calibration performance, FVIB  outperforms other IB and calibration methods by enabling continuous optimization of $\beta$. Our codes are available at \url{https://github.com/sotakudo/fvib}.

\end{abstract}

\section{Introduction}

In supervised representation learning, the fundamental objective is to extract information about a target random variable from a source random variable. Information Bottleneck (IB) \cite{tishby2000information} formalizes this from the perspective of information theory. Let $X$ be the source random variable and $Y$ be the target random variable, and assume their joint distribution is known. The goal of IB is to obtain a random variable $Z$ that is maximally informed about $Y$ while compressing the information of $X$ to some extent. Formally,
\begin{equation}
\label{ib_objective}
    \max_{Z \in \Delta} I(Z,Y) \: s.t. \: I(X, Z) \leq r 
\end{equation}
where $\Delta$ represents the set of all random variables $Z$ that satisfy the Markov chain $Y \leftrightarrow X \leftrightarrow Z$. In practice, the following IB Lagrangian \cite{gilad2003information}, a Lagrangian relaxation \cite{lemarechal2001lagrangian} of Equation (\ref{ib_objective}) is often maximized.
\begin{equation}
\label{ib_lagrangian}
    \mathcal{L}_{IB}(Z;\beta)= I(Z,Y)-\beta I(X, Z) 
\end{equation}
Notably, Variational Information Bottleneck (VIB) \cite{alemi2016deep} enables learning the IB Lagrangian, which includes intractable integrals, through a variational approximation. This has become the standard approach for IB in deep learning.
\begin{table}[t]
\caption{FVIB learns across all reasonable values of $\beta$ typically with fewer parameters and less training time compared to a single training of VIB. The values are based on the training settings for the MNIST dataset in the original VIB paper.}
\label{tab:param}
\vskip 0.15in
\begin{center}
\begin{small}
\begin{sc}
\scalebox{0.95}{
\begin{tabular}{lccc}
\toprule
\multirow{2}{*}{Methods}  &Parameters& Run-time & The number\\
  &($\times 10^5$) & (s) & of $\beta$\\
\midrule
VIB            &23.8 &528 & 1\\
FVIB (ours)    &\bf{18.6}   &  \bf{439}   & \bm{$\infty$}\\
\bottomrule
\end{tabular}
}
\end{sc}
\end{small}
\end{center}
\vskip -0.1in
\end{table}

The advantage of IB is its ability to explicitly control the trade-off between compression $I(X, Z)$ and prediction $I(Z, Y)$ through the Lagrange multiplier $\beta$. A smaller $\beta$ leads to a more predictive representation, whereas a larger $\beta$ results in a more concise one. This can also have a positive impact on deep learning models, which can sometimes be overly flexible. For example, the application of the IB framework has been theoretically or empirically shown to enhance generalization \cite{shamir2010learning, tishby2015deep, alemi2016deep, vera2018role,yu2021deep}, robustness against adversarial attacks \cite{alemi2016deep,yu2021deep, pan2021disentangled}, out-of-distribution detection \cite{alemi2018uncertainty, pan2021disentangled}, domain generalization \cite{ahuja2021invariance, li2022invariant} and calibration \cite{alemi2018uncertainty}.

However, generally, the value of  $\beta$ to be learned is not known in advance. This is because, as typically discussed by \citet{shamir2010learning}, the optimal trade-off  depends on the task-specific distribution. Therefore, practitioners are required to train multiple learners while varying $\beta$, and then follow a procedure to select the model with the best properties from these. This exploration is computationally expensive. To address this challenge in classification tasks, we propose Flexible Variational Information Bottleneck (FVIB), an innovative framework that learns the IB Lagrangian for all reasonable $\beta$ value through a singular efficient training.  FVIB can be trained using simple mean squared error, eliminating the need for external learners such as hypernetworks \cite{ha2017freedom}. Furthermore, as demonstrated in Table \ref{tab:param}, it typically outperforms even a single training process of VIB in terms of the memory and computational efficiency, while being capable of obtaining models with any reasonable value of $\beta$.

This study begins with an analysis of VIB, where we aim to find the optimal solution for an approximation of the VIB objective. Based on these results, we construct a framework. We then theoretically demonstrate that the framework is capable of learning the approximation of the VIB objective for all $\beta$ values with one training. Then empirical evaluations across various datasets and architectures confirm that FVIB can indeed learn the VIB objective for any $\beta$ as effectively as the standard VIB, but with the advantage of requiring only one training process. Finally, it is shown that continuous optimization of $\beta$ through post-processing in FVIB significantly refines calibration performance, outshining other IB methods and established calibration techniques.


\section{Analysis of Variational Information Bottleneck}
To obtain models corresponding to all $\beta$ in a single training process, we begin with an analysis of VIB. Below, we first provide an explanation of VIB, followed by an introduction to the assumptions and notations used in the analysis. Finally, we introduce the approximation of the VIB objective and derive its optimal solution.
\subsection{Variational Information Bottleneck}
VIB enables the learning of IB in deep learning by providing a lower bound of the IB Lagrangian through a variational approximation. In predicting $Y$ from $X$, a random variable Z is obtained through the feature extractor $p_{\theta}(Z|X)$. First, we consider the prediction term, the first term in Equation (\ref{ib_lagrangian}). The following variational lower bound is obtained by using a new model $q_{\phi}(Y|Z)$ as a variational approximation to $p_{\theta}(Y|Z)$.
\begin{equation}
\label{lower_pred}
\begin{split}
I(Z,Y) \ge  \int dx\: dy\: dz \: p(x,y)\: p_{\theta}(z|x)&\log q_{\phi}(y|z) \\
&+ H(Y)
\end{split}
\end{equation}
Here, $H(Y)$ represents the entropy of Y, which remains constant throughout the learning process. Next, for the compression $I(X,Z)$ in Equation (\ref{ib_lagrangian}), an upper bound can be obtained by using $r(Z)$ as a variational approximation to $p_{\theta}(Z)$.
\begin{equation}
\label{upper_comp}
I(X,Z) \le \int dx \: p(x)\: D_{KL}\sbra*{p_{\theta}(Z|x)||r(Z)}
\end{equation}
In practice, a pre-assumed distribution is used for $r(Z)$. Combining these, we obtain lower bound for the IB Lagrangian. Given the data $\cbra*{(x_1, y_1),...,(x_N, y_N)}$ and using the empirical distribution as the joint distribution of $X$ and $Y$, the objective function of VIB is derived.
\begin{equation}
\label{objective}
\begin{split}
\mathcal{L}_{VIB}(\theta, \phi;\beta) := &\frac{1}{N} \sum_{i=1}^N\left[\mathbb{E}_{p_{\theta}(Z|x_i)}\sbra*{\log q_{\phi}(y_i|Z)}\right.\\
&\left.-\beta D_{KL}\sbra*{p_{\theta}(Z|x_i)||r(Z)}\right]
\end{split}
\end{equation}

Here, we consider the case where $\beta \in [0,1]$. This is because when $\beta \ge 1$, the optimal Z becomes trivial and independent of X \cite{wu2020learnability}.

\subsection{Settings for Analysis}
\label{subsection_notation}
In the following, we introduce assumptions and notations for the model and data for our analysis. All subsequent Lemmas, Theorems, and Propositions assume the settings below. 
We consider a $d$-class classification problem with $x\in \mathcal{X}$ and $\boldsymbol{y} \in \mathbb R^d$ as one-hot vector. For the random variable $Z$, we allow $\boldsymbol{z}\in \mathbb R^{\kappa}$, where $\kappa \in \mathbb N$ is a variable, thus allowing for its optimization.
We define $p_{\theta}(Z|x)$ as follows:
\begin{equation}
p_{\theta}(Z|x)= \mathcal{N}\rbra*{\mu(x), \Sigma(x)}
\end{equation}
where $\mu:\mathcal{X}\rightarrow\mathbb{R}^{\kappa}$ and $\Sigma:\mathcal{X}\rightarrow \cbra{A \in \mathbb{R}^{\kappa \times \kappa}|A \succ 0}$. While the covariance matrix is typically limited to diagonal matrices, our analysis does not restrict it to this. Based on these settings, the parameters are $\theta = \cbra{\mu, \Sigma} $. The distribution $r(Z)$ is typically set as $r(Z)= \mathcal{N}(\mathbf{0}, I)$, and we adopt this in our study as well.
In most cases, the classifier $q_{\phi}$ consists of one dense layer and softmax function. In our analysis, the bias term is limited to $\boldsymbol{0}$ and $q_{\phi}$ is defined as
\begin{equation}
q_{\phi}(\boldsymbol{y}|\boldsymbol{z})=\boldsymbol{y}^\top \text{Softmax}(W\boldsymbol z)
\end{equation}
where $W \in \mathbb{R}^{d \times \kappa}$ is a weight matrix. 
Therefore, $\phi = \cbra{W} $. Note that during optimization of $\theta$ and $\phi$, the value of $\kappa \in \mathbb{N}$ can be varied, while ensuring its consistency between these parameters.

In addition to the above settings, we assume that the training data is class balanced. This assumption does not lose generality in practice. Because in practice, when encountering an imbalance dataset, we use undersampling \cite{cieslak2008start, peng2019trainable} or oversampling \cite{chawla2002smote} to balance the classes in the training data.


\subsection{The Optimal Solution}
Under the above settings, we analyze the VIB objective. However, it is challenging to analytically calculate the expectation term in the VIB objective. To address this, we consider an approximation of the VIB objective that can be analyzed. By using a second-order Taylor expansion of the function $f_{\boldsymbol y}(\boldsymbol z):=\log q_{\phi}(\boldsymbol y| \boldsymbol{z})$ around $\boldsymbol{z} = \boldsymbol{0}$, $T_{\boldsymbol y}(\boldsymbol z)$, which we call the Taylor approximation of log likelihood, is defined as
\begin{equation}
\label{ce_taylor_approx}
T_{\boldsymbol y}(\boldsymbol z):= f_{\boldsymbol y}(\boldsymbol 0)+\nabla f_{\boldsymbol y}(\boldsymbol 0)^{\top}\boldsymbol z+\frac{1}{2}\boldsymbol z^{\top}\nabla^2 f_{\boldsymbol y}(\boldsymbol 0)\boldsymbol z
\end{equation}
By replacing $\log q_{\phi}(y_i|Z)$ in the VIB objective with $T_{\boldsymbol y_i}(Z)$, we define the approximated objective as
\begin{equation}
\label{approx_objective}
\begin{split}
\hat{\mathcal{L}}_{VIB}( \theta, \phi;\beta) :=& \frac{1}{N} \sum_{i=1}^N\left[\mathbb{E}_{p_{\theta}(Z|x_i)}[T_{\boldsymbol y_i}(Z)]\right.\\
&\left.-\beta D_{KL}\sbra*{p_{\theta}(Z|x_i)||r(Z)}\right]
\end{split}
\end{equation}
It is named the Taylor approximation of the VIB objective here. This approximation allows us to analytically calculate the expectation term. Recalling that $r(Z)= \mathcal{N}(\mathbf{0}, I)$, the approximation error is likely to be small as $Z$ following $p_{\theta}(Z|x_i)$ is distributed near $\boldsymbol{z} = \boldsymbol{0}$ due to the KL divergence term. The detailed effect of the approximation is discussed below in Section \ref{section_ct}. We also experimentally confirm the effect of this approximation in Section \ref{subsection_taylor_results}. In the following, we consider maximizing $\hat{\mathcal{L}}_{VIB}(\phi, \theta; \beta)$.
\vspace{2mm}
\begin{lemma}
\label{prop_optimal_point}
Consider the settings in Section \ref{subsection_notation}. For any $\beta \in [0,1]$, when $\mu, \Sigma$ and $W$ satisfy the following conditions, $\hat{\mathcal{L}}_{VIB}( \theta, \phi;\beta)$ is maximized.
\newpage
\begin{itemize}
    \item $ \kappa = d-1$.
    \item For all $i = 1,2...,N$, \; $\mu(x_i) = \sqrt{1-\beta}L\rbra*{\boldsymbol{y}_i-\frac{\boldsymbol{1}}{d}}$.
    \item For all  $i = 1,2...,N$,\; $\Sigma(x_i) = \beta I$.
    \item $W=\sqrt{1-\beta}L^{\top}$.
\end{itemize}
Here, $L \in \mathbb R ^{(d-1)\times d}$ is a constant matrix that depends solely on the number of classes $d$ (see Appendix \ref{subsection_l} for details).
\end{lemma}
\vspace{2mm}
Refer to Appendix \ref{subsection_proof_lemma} for the proof.
Below, we discuss the implications of Lemma \ref{prop_optimal_point}. First, the dimension $\kappa$ of $\boldsymbol{z}$ is sufficient at $d-1$. It is also suggested that $W$ which maximizes $\hat{\mathcal{L}}_{VIB}(\phi, \theta;\beta)$ is determined independently of the training data and thus does not require learning. Similarly, $\Sigma$ does not require learning if $\Sigma(x) = \beta I$, as it becomes optimal regardless of the training data. On the other hand, it is shown that the optimal $\mu$ depends on the training data, necessitating a learning process. At $\beta=1$, optimal $p_{\theta}(\boldsymbol{z}|x)$ equals $\mathcal{N}(\boldsymbol{0},I)$, making $Z$ independent of $X$, which aligns with previous analysis of the IB Lagrangian \cite{wu2020learnability}. Therefore, the model becomes unclassifiable at $\beta = 1$. At $\beta=0$, $\Sigma(x_i) = O$ becomes optimal, making the model deterministic, consistent with previous observations \cite{alemi2016deep} in VIB. Intriguingly, although our initial model formulation did not restrict $\Sigma(x)$ to being a diagonal covariance matrix, the derived optimal solution is diagonal. This implies that in VIB, the constraint of using diagonal covariance does not affect the learning process.
\begin{figure}[t]
\vskip 0.2in
\begin{center}
\centerline{\includegraphics[width=1\columnwidth]{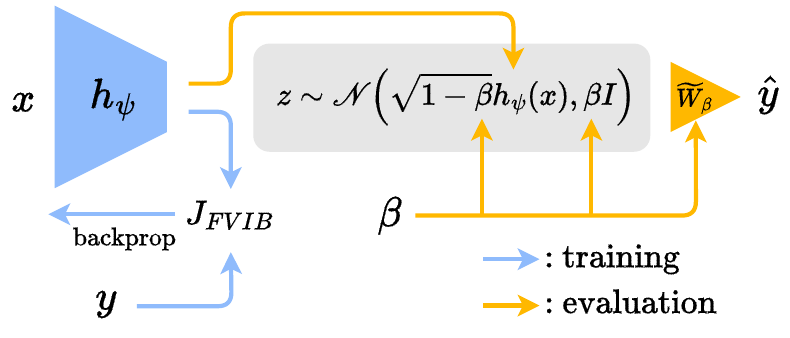}}
\caption{The $\beta$-independent training phase and $\beta$-dependent evaluation phase of FVIB.}
\label{fig:model_arch}
\end{center}
\vskip -0.2in
\end{figure}

\section{Methods}
The development of FVIB comprises two key steps: firstly, exploring a method to simultaneously maximize the Taylor approximation of the VIB objective for all $\beta \in [0,1]$; and secondly, addressing a shortcoming inherent in the Taylor approximation with a simple modification.
\subsection{Simultaneously Maximizing the Taylor Approximation of the VIB Objective for All $\beta$}
Here, we first introduce the main setup of FVIB, followed by theoretical demonstrations of its capability to maximize the Taylor approximation of the VIB objective. We train a model $h_{\psi}:\mathcal{X}\rightarrow\mathbb R^{d-1}$ with parameters $\psi$ to maximize the following objective function. It is important to note that this objective function is independent of $\beta$.
\begin{equation}
\label{our_objective}
J_{FVIB}(\psi) := -\frac{1}{N} \sum_{i=1}^N\norm*{h_{\psi}(x_i)-L\rbra*{\boldsymbol{y}_i-\frac{\boldsymbol{1}}{d}}}_2^2
\end{equation}
Next, using the trained $h_{\psi}$, for any $\beta \in [0, 1]$, we set $\mu, \Sigma$ and $W$ respectively as follows:
\begin{itemize}
    \item $\Tilde{\mu}_{\beta, \psi}(x) := \sqrt{1-\beta}h_{\psi}(x)$,
    \item $\Tilde{\Sigma}_{\beta}(x) := \beta I$,
    \item $\Tilde{W}_{\beta} :=\sqrt{1-\beta}L^{\top}$.
\end{itemize}
The flow of the variables in FVIB is summarized in Figure \ref{fig:model_arch}. Notably, the value of $\beta$ can be adjusted during the evaluation phase after the training is complete. 

We discuss below that this setup and the objective function can learn the Taylor approximation of the VIB objective for any $\beta$.
\vspace{2mm}
\begin{theorem}
\label{prop_uniform_conv}
Consider the settings in Section \ref{subsection_notation}. If $\lim_{t\rightarrow \infty}J_{FVIB}(\psi_{t})= 0 $, then the sequence of $\hat{\mathcal{L}}_{VIB}$, $\cbra*{\hat{\mathcal{L}}_{VIB}\rbra*{\Tilde{\mu}_{\beta, \psi_t}, \Tilde{\Sigma}_{\beta}, \Tilde{W}_{\beta}; \beta}}_{t \in \mathbb{N}}$ converge uniformly to $\max_{\theta, \phi}\hat{\mathcal{L}}_{VIB}(\theta, \phi; \beta)$ for $\beta \in \sbra*{0, 1}$ as $t \rightarrow \infty$.
\end{theorem} 
\vspace{2mm}
The proof can be found in Appendix \ref{subsection_proof_thm1}. It show that once $J_{FVIB}(\psi)$ is sufficiently close to the maximum, the Taylor approximation of the VIB objective approaches the maximum similarly with respect to $\beta \in [0,1]$.
This property is advantageous for our goal of simultaneously optimizing for all values of $\beta$. Theorem \ref{prop_uniform_conv} is effective when $J_{FVIB}(\psi)$ converges to zero. Although gradient descent is used and the convergence may only reach a local maximum rather than strictly zero, the following property still brings benefits regardless of the value to which $J_{FVIB}(\psi)$ converges.
\vspace{2mm}
\begin{theorem}
\label{prop_mono_inc}
Under the settings in Section \ref{subsection_notation}, an increase in the value of $J_{FVIB}(\psi)$ always leads to a higher value of $\hat{\mathcal{L}}_{VIB}\rbra*{\Tilde{\mu}_{\beta, \psi}, \Tilde{\Sigma}_{\beta}, \Tilde{W}_{\beta}; \beta}$ for any $\beta \in [0, 1) $.
\end{theorem}
\vspace{1mm}
For $\beta \in [0, 1)$, $\hat{\mathcal{L}}_{VIB}\rbra*{\Tilde{\mu}_{\beta, \psi}, \Tilde{\Sigma}_{\beta}, \Tilde{W}_{\beta}; \beta}$ can be represented as a linear function of $J_{FVIB}(\psi)$ with positive slope, thereby illustrating this characteristic. The detailed proof can be found in Appendix \ref{subsection_proof_thm1}. Note that when $\beta=1$, $\Tilde{\mu}_{\beta, \psi}(x)$ becomes $\boldsymbol 0$ regardless of $\psi$ ensuring that Taylor approximation of the VIB objective is always maximized. Theorem \ref{prop_uniform_conv} and  \ref{prop_mono_inc} indicate that learning $J_{FVIB}(\psi)$ results in the maximization of the Taylor approximation of the VIB objective, monotonically increasing it. These characteristics demonstrate that our proposed setup and objective function are effective in simultaneously maximizing the Taylor Approximation of the VIB objective for all $\beta$.

\subsection{Confidence Tuning}
\label{section_ct}
\begin{figure}[t]
\vskip 0.2in
\begin{center}
\centerline{\includegraphics[width=0.6\columnwidth]{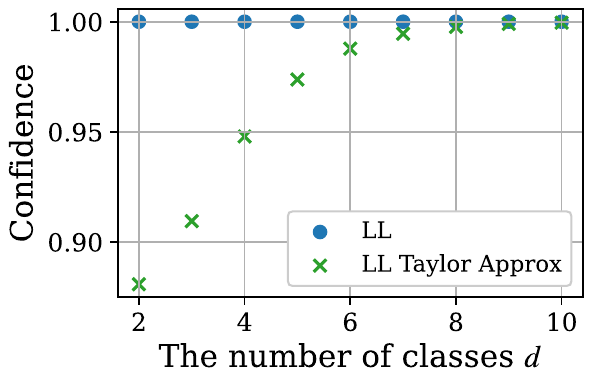}}
\caption{
Theoretical values of confidence that enables maximization of log likelihood (LL) or its Taylor approximation $T_{\boldsymbol y}(\boldsymbol{z})$.
When the number of classes $d$ is small, the restriction caused by the Taylor approximation becomes significant.}
\label{fig:opt_conf}
\end{center}
\vskip -0.2in
\end{figure}
In this section, we examine an effect of the Taylor approximation and introduce a simple adjustment to mitigate this impact. In classification, the output of a model is characterized by two aspects:  the class prediction and the confidence (i.e., the predicted probability of correctness) \cite{guo2017calibration}. The following can be said about these aspects of optimal models for the Taylor approximation of log likelihood.

\begin{proposition}
\label{prop_opt_conf}
When $T_{\boldsymbol y}(\boldsymbol{z})$ is maximized under the settings in Section \ref{subsection_notation}, the class prediction matches the label and the confidence is $\frac{\exp(d)}{\exp(d)+d-1}$.
\end{proposition}
The proof can be found in Appendix \ref{subsection_proof_prop}.
On the other hand, original log likelihood is maximized when the class prediction matches the label and the confidence equals $1$. These facts indicate that applying Taylor approximation doesn't change the class prediction while it limits the confidence. As demonstrated in Figure \ref{fig:opt_conf}, especially when the number of classes $d$ is small, the restriction becomes significant. When the class prediction is correct, the log likelihood value in $\mathcal{L}_{VIB}$ increases as confidence approaches $1$. Therefore, the restriction is counterproductive for maximizing the VIB objective, conflicting with our aim. To counteract this limitation, we propose a straightforward adjustment to the confidence while leaving the class prediction unchanged. We adopt Temperature Scaling \cite{hinton2015distilling}, which divides the logit by some temperature $T$.
In our adjustment, $T$ is defined as the temperature at which the optimal model for the Taylor approximation of log likelihood achieves a confidence level of $c$, which is sufficiently close to $1$. $T$ is thus given by $T=\frac{d}{\log \frac{(d-1)c}{1-c}}$.
It ensures that at $\beta=0$ , the optimal model of the Taylor approximation of the VIB objective achieves the confidence level $c$ in the training data. For the sake of simplicity, we ignore the impact of $\beta$ and consistently use the same temperature. This approach, which we term Confidence Tuning (CT) requires no additional training. It solely modify the variational distribution $q_{\phi}$ while keeping the representation $Z$ unchanged. In this study, we consistently set $c=0.997$ in all experiments. Note that, if we want to tune $c$, it can also be done after the training.


\section{Related works}
\subsection{Efficient Sweep of $\beta$ in IB}
\citet{wu2020learnability} demonstrated that, even for $\beta \in [0,1)$, inappropriate values of $\beta$ can make IB Lagrangian unlearnable (i.e., making the optimal $Z$ independent of $X$). They theoretically established sufficient conditions for $\beta$ to be learnable. Based on this, they proposed an algorithm to estimate the range of learnable $\beta$. This research is valuable from a search efficiency perspective, as it narrows down the range of $\beta$ to be investigated. However, the necessity of multiple trainings to find a useful $\beta$ still remains.
\citet{rodriguez2020convex} developed a method that realizes the desired compression rate, $r$ in Equation (\ref{ib_objective}), in a single training by providing a bijection between the Lagrange multiplier $\beta$ and the compression rate. However, trial and error is ultimately necessary to identify a suitable compression rate, and this challenge persists. Moreover, \citet{pan2021disentangled} enabled the acquisition of a maximally compressed representation $Z$ without reducing $I(Z, Y)$ in the training data using supervised disentangling. While this learning process is independent of $\beta$, there is no guarantee that this representation is the most useful, as our aim is often to maximize the true $I(Z, Y)$, not the empirical $I(Z, Y)$ in the training data. In fact, similar to the bias-variance trade-off, the true $I(Z, Y)$ is determined by a trade-off between bias (here, empirical $I(Z, Y)$ in the training data) and variance, which is controlled by $I(X,Z)$ \cite{shamir2010learning}. Since this trade-off depends on the task-specific distribution, designing a fixed loss that always leads to the optimal trade-off is challenging. Therefore, we adopt a different approach, designing a model that learns all $\beta$ with single training, significantly reducing the search cost.
\subsection{Variational Autoencoders}
Variational Autoencoder (VAE) \cite{kingma2013auto} and $\beta$-VAE \cite{higgins2016beta} have a loss with both distortion (prediction) and rate (compression) terms, and can be interpreted as special cases of VIB \cite{alemi2016deep, alemi2018fixing, tschannen2018recent}. This research is related to several studies on VAEs, owing to this similarity. MR-VAE \cite{bae2022multi} enables the creation of models in $\beta$-VAE that can be modified for any $\beta$ after training. This is made possible by learning a response function that transforms $\beta$ into optimal parameters using a hypernetwork. In contrast, our study, through the detailed analysis of VIB, enables the creation of flexible models for classification task simply by training with mean squared error, without the need for any external learner. Additionally, numerous analyses have been conducted to understand the optimal solution properties of $\beta$-VAE \cite{lucas2019don, kumar2020implicit, sicks2021generalised}. Our analysis reconsiders these in the setting of VIB.

\section{Experiments}
Our experiments initially address the following three key questions: (1) Can the Taylor approximation of the VIB objective  effectively learn the VIB objective?; (2) Does FVIB successfully maximize the Taylor approximation of the VIB objective?; (3) Is FVIB capable of effectively learning the VIB objective? Following these investigations, we present the results of FVIB in terms of calibration and generalization performance. For details about the model architectures and the training settings, please refer to Appendix \ref{subsection_setup}.
\subsection{Can the Taylor Approximation of the VIB Objective Effectively Learn the VIB Objective?}
\label{subsection_taylor_results}
\begin{figure}[t]
\vskip 0.2in
\begin{center}
\centerline{\includegraphics[width=\linewidth]{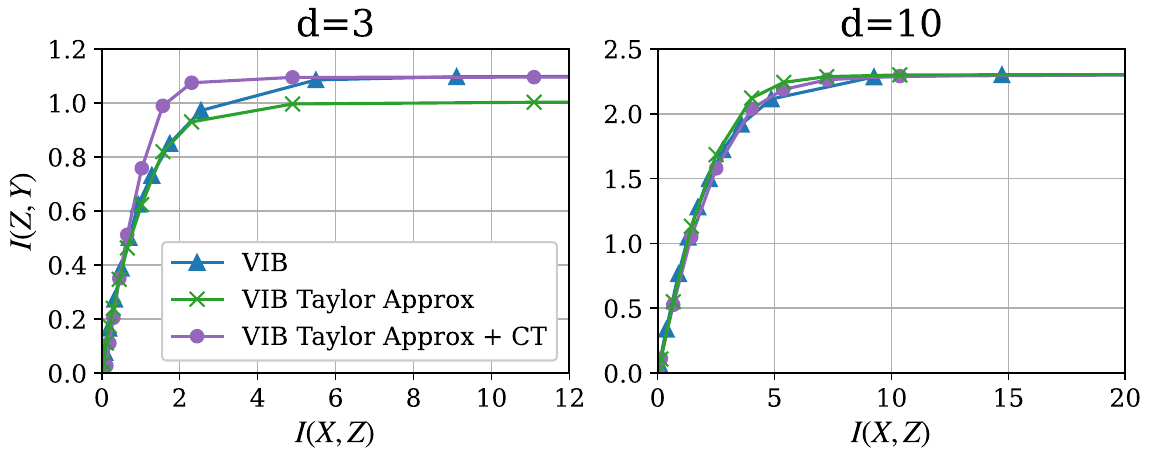}}
\caption{IB curves for the training data obtained by VIB or its Taylor approximation with or without CT. For the values of $I(X, Z)$ and $I(Z, Y)$, R.H.S of Equation (\ref{upper_comp}) and Equation (\ref{lower_pred}) are used. The Taylor approximation with CT effectively learn the VIB objective.}
\label{vib_vs_taylor}
\end{center}
\vskip -0.2in
\end{figure}

\begin{figure*}[h]
\vskip 0.2in
\begin{center}
\centerline{\includegraphics[width=0.7\linewidth]{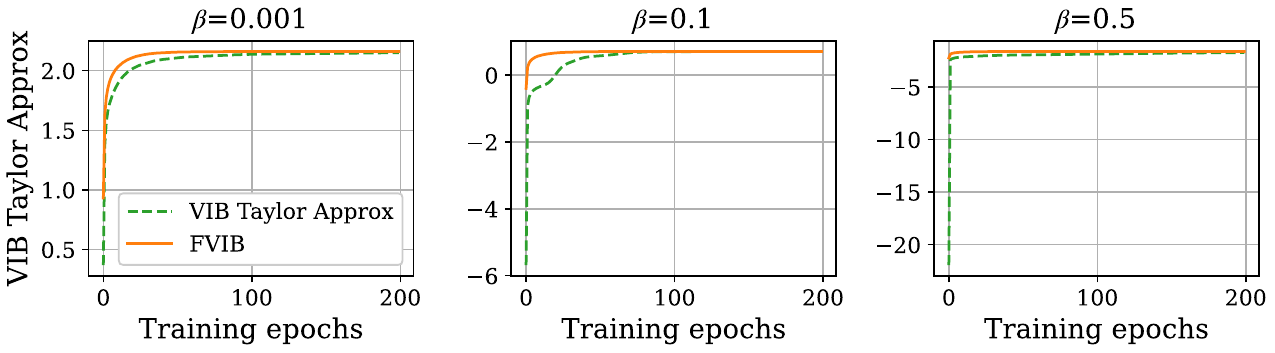}}
\caption{Comparison of the Taylor approximation values of the VIB objective during training, both in the direct training and when using FVIB. FVIB enables effective optimization for each $\beta$ while it needs only one training.}
\label{fig:FVIB_opt_VIB_Taylor}
\end{center}
\vskip -0.2in
\end{figure*}

\begin{figure*}[ht]
\vskip 0.2in
\begin{center}
\centerline{\includegraphics[width=\linewidth]{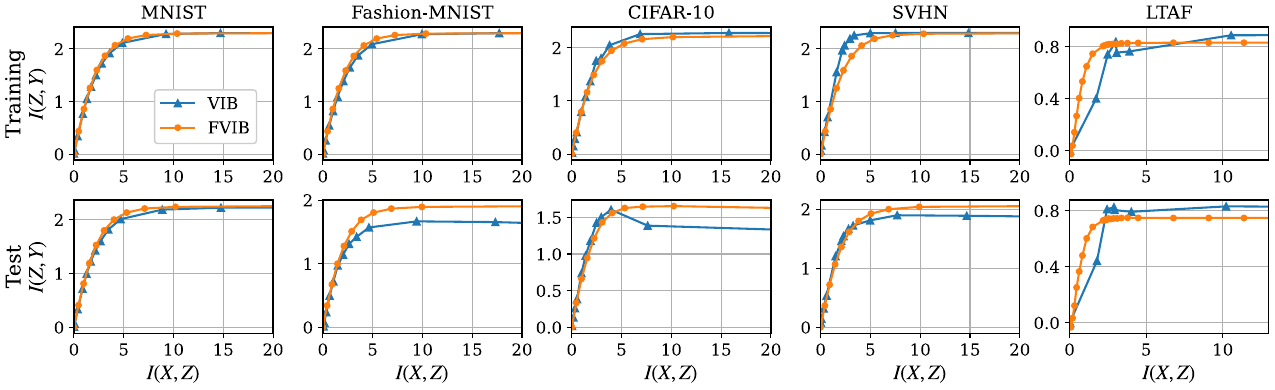}}
\caption{Comparison of IB curves obtained by FVIB and VIB. FVIB requires only a single training to construct the IB Curve, while VIB necessitates multiple trainings. FVIB achieves IB curves that are competitive with those produced by VIB.}
\label{fig:FVIB_IB_curve}
\end{center}
\vskip -0.2in
\end{figure*}

FVIB is designed to learn the Taylor approximation of the VIB objective. Therefore it is crucial to verify whether learning the Taylor approximation of the VIB objective can effectively learn the VIB objective. In this part, we train models using the VIB objective or its Taylor approximation, systematically varying the $\beta$.  By plotting the compression term, the R.H.S of Equation (\ref{upper_comp}) versus the prediction term, the R.H.S of Equation (\ref{lower_pred}) we create the IB curves for the two objectives. We use two datasets: the entire MNIST dataset \cite{larochelle2011neural} and its subset containing only the first three classes (i.e., digits 0, 1, 2). Training is conducted with $\beta \in \{10^{-6}, \dots, 10^{-2}, 0.1, 0.2, \dots, 1.0\}$. The model consists of four fully connected layers with a stochastic embedding before the final layer.
In Figure \ref{vib_vs_taylor}, we present the IB curves obtained from the VIB objective, and its Taylor approximation, both with and without CT, in the training data. For $d=10$, the IB curves obtained from the Taylor approximation of the VIB objective, both with and without CT, align well with those obtained from the VIB objective. On the other hand, for $d=3$, as discussed earlier, its Taylor approximation tends to have a lower prediction term value due to the limited confidence, compared to the VIB objective. The gap is shortened by adjusting the confidence through CT, resulting in equal or higher prediction term values compared to the VIB objective. Overall, our experiments demonstrate that its Taylor approximation with CT effectively maximizes the VIB objective, often outperforming the standard approach. The same trend is observed in the test data. Note that the representation $Z$ remains unchanged with or without CT. The observed effect of CT is thus not due to an increase in $I(Z, Y)$ itself, but rather due to the variational bound of Equation (\ref{lower_pred}) becoming tighter as the classifier $q_{\phi}$ is optimized further.

\subsection{Does FVIB Successfully Maximize the Taylor Approximation of the VIB Objective?}
The next step is to verify whether FVIB can effectively maximize the Taylor approximation of the VIB objective. To this end, we conduct training using both the Taylor approximation of the VIB objective and FVIB on the entire MNIST dataset. Figure \ref{fig:FVIB_opt_VIB_Taylor} shows the values of the Taylor approximation of the VIB objective using the training data across different training epochs, for $\beta \in \cbra*{0.001, 0.1, 0.5}$. Note that while the learning of the Taylor approximation of the VIB objective is conducted separately for each $\beta$, the training of FVIB is performed only once. For each $\beta$, the values for FVIB monotonically increases and converges to a value similar to that achieved when trained with the Taylor approximation of the VIB objective itself. These results are consistent with the theoretical properties of FVIB. Moreover, our results indicate that FVIB enables faster convergence for each $\beta$ value compared to the direct training.

\begin{figure*}[ht]
\vskip 0.2in
\begin{center}
\centerline{\includegraphics[width=0.8\linewidth]{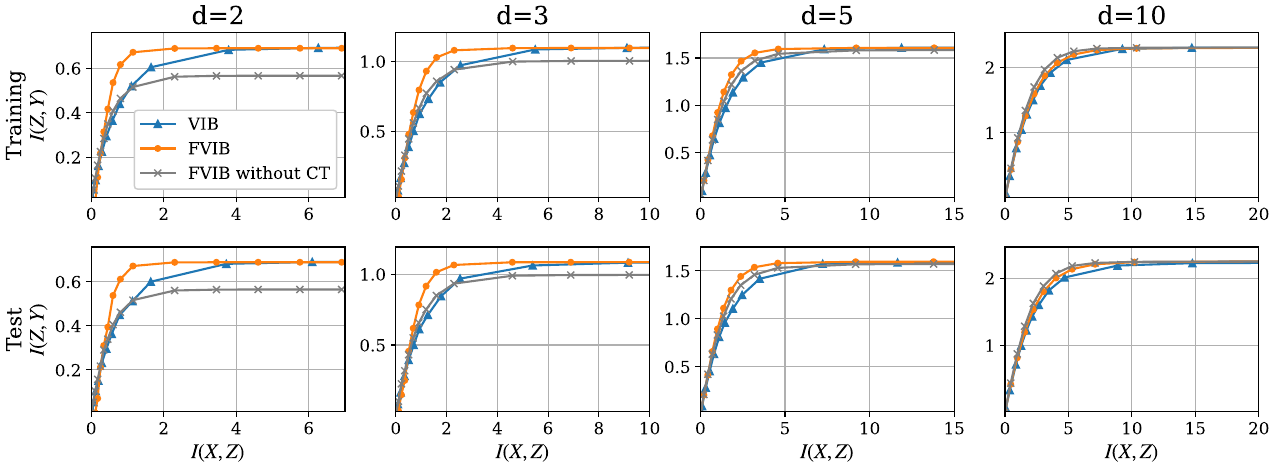}}
\caption{IB Curves obtained by VIB and FVIB with and without CT, using subsets of the MNIST dataset with varying numbers of classes. When the number of classes is small, the values of the prediction term for FVIB without CT are limited, and CT effectively recovers these values.}
\label{fig:FVIB_IB_curve_abration}
\end{center}
\vskip -0.2in
\end{figure*}

\subsection{Is FVIB Capable of Effectively Learning the VIB Objective?}
To compare the IB curves obtained by FVIB and VIB, we use various datasets, including the image datasets MNIST \cite{larochelle2011neural}, Fashion-MNIST \cite{xiao2017fashion}, CIFAR-10 \cite{krizhevsky2009learning}, SVHN \cite{netzer2011reading}, and the time series dataset Long Term AF (LTAF) \cite{petrutiu2007abrupt}. Particularly, LTAF is preprocessed based on \citet{kudo2023training} for a three-class classification of arrhythmia from ECG data. For MNIST and Fashion-MNIST, models with fully connected layers are used; for CIFAR-10 and SVHN, CNN-based models; and for LTAF, LSTM-based models are employed (details in Table \ref{tab:encoder} in Appendix). Training is conducted with $\beta \in \{10^{-6}, ..., 10^{-2}, 0.1, 0.2, ... 1.0\}$.  For the LTAF dataset, extra values $\beta \in \{0.02, 0.03... 0.09\}$ is also included. The respective IB curves are shown in Figure \ref{fig:FVIB_IB_curve}. Note that, unlike VIB, which requires multiple training processes, FVIB is trained only once for each dataset. Despite this streamlined training process, the performance of FVIB aligns closely with that of the traditional VIB in the training data. Moreover, in the test data, FVIB attains higher values of prediction term than VIB for all datasets except  the LTAF. These results demonstrate the effectiveness and robustness of the representations obtained by FVIB.

Next, to assess the necessity of CT, we examine how the absence of CT affects the IB curves for FVIB while varying the number of classes $d$. Similar to the previous experiments, we use the subsets of MNIST for each $d \in \{2,3,5,10\}$.  As shown in Figure \ref{fig:FVIB_IB_curve_abration}, in every scenario, the IB curves generated by FVIB tend to exceed those produced by VIB. Particularly, when $d$ is small, the absence of CT results in limited confidence, leading to smaller prediction terms, indicating the efficacy of the CT. Figure \ref{fig:FVIB_IB_curve} and Figure \ref{fig:FVIB_IB_curve_abration} show that FVIB, despite being trained only once, can learn the VIB objective without compromising their quality.

\begin{table*}[h]
\caption{ECE (\%) (lower is better) for FVIB compared to calibration methods and other IB methods. An asterisk (*) indicates optimization of $\beta$ through model selection, while a dagger ($\dag$) signifies the continuous optimization of $\beta$. FVIB with continuous optimization achieves the best average ECE.}
\label{tab:calibration}
\vskip 0.15in
\begin{center}
\begin{small}
\begin{sc}
\begin{tabular}{lcccccc}
\toprule
Methods &MNIST &Fashion-MNIST&CIFAR-10&SVHN&LTAF&Avg \\
\midrule
Baseline               & 1.20     &8.72      & 10.66   &7.06     & 2.56 &6.04\\
\midrule
TS    &0.46      & \bf{1.03}& 1.97& 1.44   &3.20 &1.62 \\
ETS    & 0.46     & 1.13& 1.83& 1.25   &2.70 &1.47 \\
IRM    &  \bf{0.24}    & 1.32& \bf{1.38}& 1.40   & 1.99 &1.27 \\
\midrule
VIB*                    & 0.58&2.36&2.58&2.25& 0.82 &1.72\\
sq-VIB*                 & 0.49&1.49&4.93&1.20     & 1.17&1.86\\
NIB*                    & 0.34     &2.07      &5.42     &2.96     & 1.14&2.39 \\
sq-NIB*                 & 1.54     &4.88      &10.88    &5.51     & 1.04&4.77\\
\midrule
FVIB ($\beta=0$)       &0.28& 5.36&9.29&2.09& 5.11 &4.43\\
FVIB*                   &0.31&2.30&3.70&\bf{0.50}&2.57&1.88\\
FVIB\dag                   & 0.29&1.78&1.77&0.56& \bf{0.72}& \bf{1.02}\\
\bottomrule
\end{tabular}
\end{sc}
\end{small}
\end{center}
\vskip -0.1in
\end{table*}


\begin{table*}[ht]
\caption{Accuracy (\%) for FVIB compared to other IB methods. FVIB surpasses baseline and other IB methods even when $\beta = 0$.}
\label{gene_tab}
\vskip 0.15in
\begin{center}
\begin{small}
\begin{sc}
\begin{tabular}{lcccccc}
\toprule
Methods &MNIST &Fashion-MNIST&CIFAR-10&SVHN&LTAF&Avg \\
\midrule
Baseline        &98.52  & 90.29&87.94&92.13& 91.81&92.14\\
\midrule
VIB             & \bf{98.74}&90.28&87.19&93.34& 91.57 &92.22\\
sq-VIB             & 98.61&90.16&87.35&93.58&91.46&92.23\\
NIB                    & 97.94     &88.69      &87.61     &93.79     & 91.36 &91.88\\
sq-NIB                 & 97.91     &88.39      &\bf{87.95}    &93.53     &91.41 &91.84\\
\midrule
FVIB ($\beta=0$) &98.73& 90.28&86.53&94.16&92.06 &92.35\\
FVIB &98.73&\bf{90.35}&86.55&\bf{94.20}&\bf{92.07}& \bf{92.38}\\
\bottomrule
\end{tabular}
\end{sc}
\end{small}
\end{center}
\vskip -0.1in
\end{table*}


\subsection{Calibration Performance of FVIB}
It has been theoretically demonstrated that compression of representation through IB contributes to the increase in the true $I(Z,Y)$ \cite{shamir2010learning}.
Considering that log likelihood reflects $I(Z, Y)$ as shown in Equation (\ref{lower_pred}), this explains why compression in IB can lead to improved log likelihood values in test data as frequently observed (e.g., CIFAR-10 test data in Figure \ref{fig:FVIB_IB_curve}). 
This suggests that IB can be useful for model calibration, and indeed, the effectiveness of VIB in calibration has been shown \cite{alemi2018uncertainty}. However, traditional IB methods set $\beta$ prior to learning, which not only necessitates multiple training processes but also prevents the continuous optimization of $\beta$. FVIB resolves this issue by optimizing $\beta$ through post-processing, paving the way for the application of IB in calibration.

We evaluate the Expected Calibration Error (ECE) \cite{naeini2015obtaining} for various IB and calibration methods, and the results are shown in Table \ref{tab:calibration}. For each dataset, the first 1,000 data points from the test set (for the LTAF dataset, 4,217 data points corresponding to three subjects to avoid subject leaks) are separated as a validation set for calibration. FVIB continuously optimizes parameter $\beta$ to minimize negative log likelihood loss in the validation set. Optimization is done using the L-BFGS method \cite{liu1989limited} with a learning rate of 0.1 and a maximum of 50 iterations in PyTorch \cite{paszke2019pytorch}. The baseline is a model trained using cross-entropy. For calibration methods, we use the standard Temperature Scaling (TS) \cite{guo2017calibration}, and more sophisticated methods such as the parametric Ensemble Temperature Scaling (ETS) \cite{zhang2020mix} and the non-parametric multi-class isotonic regression (IRM) \cite{zhang2020mix}. TS and ETS are trained to minimize negative log likelihood loss in validation data. All training is based on the code from \cite{zhang2020mix}.  For IB methods, we compared VIB, Nonlinear Information Bottleneck (NIB) \cite{kolchinsky2019nonlinear}, sq-VIB and sq-NIB \cite{kolchinsky2018caveats}, which use a squared compression term in the IB Lagrangian. These are trained with $\beta \in \{10^{-6}, ..., 10^{-2}, 0.1, 0.2, ... 1.0\}$, selecting $\beta$ that minimize ECE in the validation set. To avoid selecting models with significantly low accuracy, candidate $\beta$ values are those with over 50\% accuracy on validation data. As in previous studies \cite{alemi2016deep, alemi2018uncertainty}, IB methods, including FVIB, sample $Z$ multiple times and average the probability for likelihood calculation. Therefore, output probability $p$ is calculated as $p(\boldsymbol{y}|x)= \frac{1}{S}\Sigma_{s=1}^{S}q_{\phi}(\boldsymbol{y}|\boldsymbol{z}^s),\; where\; \boldsymbol{z}^s \sim p_{\theta}(\boldsymbol{z}|x)$. In this experiment, the number of samples $S$ is set to $30$. Note that since $Z$ is sampled before the final layer, the increase in computational cost are limited.

FVIB, like other IB methods, improves calibration performance through compression of representation. Furthermore, in FVIB, the continuous optimization of $\beta$ via post-processing leads to better calibration compared to optimization through discrete hyperparameter search. 
This result suggests the usefulness of continuous optimization of $\beta$ in calibration. Consequently, FVIB demonstrates superior calibration performance compared to other IB and calibration methods in the average of five datasets.

\subsection{Generalization Performance of FVIB}
Finally, we compare the accuracy of FVIB with the other IB methods. The results are shown in Table \ref{gene_tab}. For IB methods, we explore $\beta \in \{10^{-6}, ..., 10^{-2}, 0.1, 0.2, ... 1.0\}$ and display the best results obtained. Here, $S$, the number of samples for $Z$ is set to $1$. We observe little change in accuracy due to compression in FVIB. However, FVIB tends to have better accuracy than the baseline even at $\beta=0$. This suggests that the Taylor approximation of log likelihood improves accuracy. This can be due to the Taylor approximation acting as a form of regularization by limiting confidence, similar to label smoothing \cite{szegedy2016rethinking} or confidence penalty \cite{pereyra2017regularizing}.
As a result of this regularization, FVIB surpass not only the baseline, but also other IB methods in terms of average accuracy across five datasets.

\section{Conclusions}

In IB, achieving the optimal compression-prediction trade-off traditionally requires multiple trainings to search for the Lagrange multiplier $\beta$. To address this, our study introduces FVIB, a framework for classification problems that disentangle the learning process of  IB from the optimization of $\beta$. We theoretically demonstrate that FVIB can learn the approximation of the VIB objective for all $\beta$ simultaneously. Then it is empirically shown that FVIB can do this without compromising its ability to learn the VIB objective. Furthermore, experiments on calibration performance show the benefit of the continuous optimization of $\beta$ enabled by FVIB. These results show that FVIB greatly advances the challenge of optimizing the trade-off and enables more flexible applications of IB.


\bibliography{example_paper}

\begin{thebibliography}{46}
\providecommand{\natexlab}[1]{#1}
\providecommand{\url}[1]{\texttt{#1}}
\expandafter\ifx\csname urlstyle\endcsname\relax
  \providecommand{\doi}[1]{doi: #1}\else
  \providecommand{\doi}{doi: \begingroup \urlstyle{rm}\Url}\fi

\bibitem[Achille \& Soatto(2018)Achille and Soatto]{achille2018information}
Achille, A. and Soatto, S.
\newblock Information dropout: Learning optimal representations through noisy computation.
\newblock \emph{IEEE transactions on pattern analysis and machine intelligence}, 40\penalty0 (12):\penalty0 2897--2905, 2018.

\bibitem[Ahuja et~al.(2021)Ahuja, Caballero, Zhang, Gagnon-Audet, Bengio, Mitliagkas, and Rish]{ahuja2021invariance}
Ahuja, K., Caballero, E., Zhang, D., Gagnon-Audet, J.-C., Bengio, Y., Mitliagkas, I., and Rish, I.
\newblock Invariance principle meets information bottleneck for out-of-distribution generalization.
\newblock \emph{Advances in Neural Information Processing Systems}, 34:\penalty0 3438--3450, 2021.

\bibitem[Alemi et~al.(2018{\natexlab{a}})Alemi, Poole, Fischer, Dillon, Saurous, and Murphy]{alemi2018fixing}
Alemi, A., Poole, B., Fischer, I., Dillon, J., Saurous, R.~A., and Murphy, K.
\newblock Fixing a broken elbo.
\newblock In \emph{International conference on machine learning}, pp.\  159--168. PMLR, 2018{\natexlab{a}}.

\bibitem[Alemi et~al.(2016)Alemi, Fischer, Dillon, and Murphy]{alemi2016deep}
Alemi, A.~A., Fischer, I., Dillon, J.~V., and Murphy, K.
\newblock Deep variational information bottleneck.
\newblock \emph{arXiv preprint arXiv:1612.00410}, 2016.

\bibitem[Alemi et~al.(2018{\natexlab{b}})Alemi, Fischer, and Dillon]{alemi2018uncertainty}
Alemi, A.~A., Fischer, I., and Dillon, J.~V.
\newblock Uncertainty in the variational information bottleneck.
\newblock \emph{arXiv preprint arXiv:1807.00906}, 2018{\natexlab{b}}.

\bibitem[Bae et~al.(2022)Bae, Zhang, Ruan, Wang, Hasegawa, Ba, and Grosse]{bae2022multi}
Bae, J., Zhang, M.~R., Ruan, M., Wang, E., Hasegawa, S., Ba, J., and Grosse, R.
\newblock Multi-rate vae: Train once, get the full rate-distortion curve.
\newblock \emph{arXiv preprint arXiv:2212.03905}, 2022.

\bibitem[Chawla et~al.(2002)Chawla, Bowyer, Hall, and Kegelmeyer]{chawla2002smote}
Chawla, N.~V., Bowyer, K.~W., Hall, L.~O., and Kegelmeyer, W.~P.
\newblock Smote: synthetic minority over-sampling technique.
\newblock \emph{Journal of artificial intelligence research}, 16:\penalty0 321--357, 2002.

\bibitem[Cieslak \& Chawla(2008)Cieslak and Chawla]{cieslak2008start}
Cieslak, D.~A. and Chawla, N.~V.
\newblock Start globally, optimize locally, predict globally: Improving performance on imbalanced data.
\newblock In \emph{2008 Eighth IEEE International Conference on Data Mining}, pp.\  143--152. IEEE, 2008.

\bibitem[Faust et~al.(2018)Faust, Shenfield, Kareem, San, Fujita, and Acharya]{faust2018automated}
Faust, O., Shenfield, A., Kareem, M., San, T.~R., Fujita, H., and Acharya, U.~R.
\newblock Automated detection of atrial fibrillation using long short-term memory network with rr interval signals.
\newblock \emph{Computers in biology and medicine}, 102:\penalty0 327--335, 2018.

\bibitem[Gilad-Bachrach et~al.(2003)Gilad-Bachrach, Navot, and Tishby]{gilad2003information}
Gilad-Bachrach, R., Navot, A., and Tishby, N.
\newblock An information theoretic tradeoff between complexity and accuracy.
\newblock In \emph{Learning Theory and Kernel Machines: 16th Annual Conference on Learning Theory and 7th Kernel Workshop, COLT/Kernel 2003, Washington, DC, USA, August 24-27, 2003. Proceedings}, pp.\  595--609. Springer, 2003.

\bibitem[Guo et~al.(2017)Guo, Pleiss, Sun, and Weinberger]{guo2017calibration}
Guo, C., Pleiss, G., Sun, Y., and Weinberger, K.~Q.
\newblock On calibration of modern neural networks.
\newblock In \emph{International conference on machine learning}, pp.\  1321--1330. PMLR, 2017.

\bibitem[Ha \& Phuong(2017)Ha and Phuong]{ha2017freedom}
Ha, D.~T. and Phuong, D.~L.
\newblock Freedom of information law comes to vietnam: How do human rights adapt to goals of economic development and political stability?
\newblock \emph{Austl. J. Asian L.}, 18:\penalty0 167, 2017.

\bibitem[Higgins et~al.(2016)Higgins, Matthey, Pal, Burgess, Glorot, Botvinick, Mohamed, and Lerchner]{higgins2016beta}
Higgins, I., Matthey, L., Pal, A., Burgess, C., Glorot, X., Botvinick, M., Mohamed, S., and Lerchner, A.
\newblock beta-vae: Learning basic visual concepts with a constrained variational framework.
\newblock In \emph{International conference on learning representations}, 2016.

\bibitem[Hinton et~al.(2015)Hinton, Vinyals, and Dean]{hinton2015distilling}
Hinton, G., Vinyals, O., and Dean, J.
\newblock Distilling the knowledge in a neural network.
\newblock \emph{arXiv preprint arXiv:1503.02531}, 2015.

\bibitem[Kingma \& Ba(2014)Kingma and Ba]{kingma2014adam}
Kingma, D.~P. and Ba, J.
\newblock Adam: A method for stochastic optimization.
\newblock \emph{arXiv preprint arXiv:1412.6980}, 2014.

\bibitem[Kingma \& Welling(2013)Kingma and Welling]{kingma2013auto}
Kingma, D.~P. and Welling, M.
\newblock Auto-encoding variational bayes.
\newblock \emph{arXiv preprint arXiv:1312.6114}, 2013.

\bibitem[Kolchinsky et~al.(2018)Kolchinsky, Tracey, and Van~Kuyk]{kolchinsky2018caveats}
Kolchinsky, A., Tracey, B.~D., and Van~Kuyk, S.
\newblock Caveats for information bottleneck in deterministic scenarios.
\newblock \emph{arXiv preprint arXiv:1808.07593}, 2018.

\bibitem[Kolchinsky et~al.(2019)Kolchinsky, Tracey, and Wolpert]{kolchinsky2019nonlinear}
Kolchinsky, A., Tracey, B.~D., and Wolpert, D.~H.
\newblock Nonlinear information bottleneck.
\newblock \emph{Entropy}, 21\penalty0 (12):\penalty0 1181, 2019.

\bibitem[Krizhevsky et~al.(2009)Krizhevsky, Hinton, et~al.]{krizhevsky2009learning}
Krizhevsky, A., Hinton, G., et~al.
\newblock Learning multiple layers of features from tiny images.
\newblock 2009.

\bibitem[Kudo et~al.(2023)Kudo, Chen, Zhou, Izu, Chen-Izu, Zhu, Tamura, Kanaya, and Huang]{kudo2023training}
Kudo, S., Chen, Z., Zhou, X., Izu, L.~T., Chen-Izu, Y., Zhu, X., Tamura, T., Kanaya, S., and Huang, M.
\newblock A training pipeline of an arrhythmia classifier for atrial fibrillation detection using photoplethysmography signal.
\newblock \emph{Frontiers in Physiology}, 14:\penalty0 2, 2023.

\bibitem[Kumar \& Poole(2020)Kumar and Poole]{kumar2020implicit}
Kumar, A. and Poole, B.
\newblock On implicit regularization in $\beta$-vaes.
\newblock In \emph{International Conference on Machine Learning}, pp.\  5480--5490. PMLR, 2020.

\bibitem[Larochelle \& Murray(2011)Larochelle and Murray]{larochelle2011neural}
Larochelle, H. and Murray, I.
\newblock The neural autoregressive distribution estimator.
\newblock In \emph{Proceedings of the fourteenth international conference on artificial intelligence and statistics}, pp.\  29--37. JMLR Workshop and Conference Proceedings, 2011.

\bibitem[Lemar{\'e}chal(2001)]{lemarechal2001lagrangian}
Lemar{\'e}chal, C.
\newblock Lagrangian relaxation.
\newblock \emph{Computational combinatorial optimization: optimal or provably near-optimal solutions}, pp.\  112--156, 2001.

\bibitem[Li et~al.(2022)Li, Shen, Wang, Zhu, Li, Keutzer, and Zhao]{li2022invariant}
Li, B., Shen, Y., Wang, Y., Zhu, W., Li, D., Keutzer, K., and Zhao, H.
\newblock Invariant information bottleneck for domain generalization.
\newblock In \emph{Proceedings of the AAAI Conference on Artificial Intelligence}, volume~36, pp.\  7399--7407, 2022.

\bibitem[Liu \& Nocedal(1989)Liu and Nocedal]{liu1989limited}
Liu, D.~C. and Nocedal, J.
\newblock On the limited memory bfgs method for large scale optimization.
\newblock \emph{Mathematical programming}, 45\penalty0 (1-3):\penalty0 503--528, 1989.

\bibitem[Lucas et~al.(2019)Lucas, Tucker, Grosse, and Norouzi]{lucas2019don}
Lucas, J., Tucker, G., Grosse, R.~B., and Norouzi, M.
\newblock Don't blame the elbo! a linear vae perspective on posterior collapse.
\newblock \emph{Advances in Neural Information Processing Systems}, 32, 2019.

\bibitem[Naeini et~al.(2015)Naeini, Cooper, and Hauskrecht]{naeini2015obtaining}
Naeini, M.~P., Cooper, G., and Hauskrecht, M.
\newblock Obtaining well calibrated probabilities using bayesian binning.
\newblock In \emph{Proceedings of the AAAI conference on artificial intelligence}, volume~29, 2015.

\bibitem[Netzer et~al.(2011)Netzer, Wang, Coates, Bissacco, Wu, and Ng]{netzer2011reading}
Netzer, Y., Wang, T., Coates, A., Bissacco, A., Wu, B., and Ng, A.~Y.
\newblock Reading digits in natural images with unsupervised feature learning.
\newblock 2011.

\bibitem[Pan et~al.(2021)Pan, Niu, Zhang, and Zhang]{pan2021disentangled}
Pan, Z., Niu, L., Zhang, J., and Zhang, L.
\newblock Disentangled information bottleneck.
\newblock In \emph{Proceedings of the AAAI Conference on Artificial Intelligence}, volume~35, pp.\  9285--9293, 2021.

\bibitem[Paszke et~al.(2019)Paszke, Gross, Massa, Lerer, Bradbury, Chanan, Killeen, Lin, Gimelshein, Antiga, et~al.]{paszke2019pytorch}
Paszke, A., Gross, S., Massa, F., Lerer, A., Bradbury, J., Chanan, G., Killeen, T., Lin, Z., Gimelshein, N., Antiga, L., et~al.
\newblock Pytorch: An imperative style, high-performance deep learning library.
\newblock \emph{Advances in neural information processing systems}, 32, 2019.

\bibitem[Peng et~al.(2019)Peng, Zhang, Xing, Gui, Huang, Jiang, Ding, and Chen]{peng2019trainable}
Peng, M., Zhang, Q., Xing, X., Gui, T., Huang, X., Jiang, Y.-G., Ding, K., and Chen, Z.
\newblock Trainable undersampling for class-imbalance learning.
\newblock In \emph{Proceedings of the AAAI conference on artificial intelligence}, volume~33, pp.\  4707--4714, 2019.

\bibitem[Pereyra et~al.(2017)Pereyra, Tucker, Chorowski, Kaiser, and Hinton]{pereyra2017regularizing}
Pereyra, G., Tucker, G., Chorowski, J., Kaiser, {\L}., and Hinton, G.
\newblock Regularizing neural networks by penalizing confident output distributions.
\newblock \emph{arXiv preprint arXiv:1701.06548}, 2017.

\bibitem[Petrutiu et~al.(2007)Petrutiu, Sahakian, and Swiryn]{petrutiu2007abrupt}
Petrutiu, S., Sahakian, A.~V., and Swiryn, S.
\newblock Abrupt changes in fibrillatory wave characteristics at the termination of paroxysmal atrial fibrillation in humans.
\newblock \emph{Europace}, 9\penalty0 (7):\penalty0 466--470, 2007.

\bibitem[Peyhardi et~al.(2014)Peyhardi, Trottier, and Gu{\'e}don]{peyhardi2014new}
Peyhardi, J., Trottier, C., and Gu{\'e}don, Y.
\newblock A new specification of generalized linear models for categorical data.
\newblock \emph{arXiv preprint arXiv:1404.7331}, 2014.

\bibitem[Rodr{\'\i}guez~G{\'a}lvez et~al.(2020)Rodr{\'\i}guez~G{\'a}lvez, Thobaben, and Skoglund]{rodriguez2020convex}
Rodr{\'\i}guez~G{\'a}lvez, B., Thobaben, R., and Skoglund, M.
\newblock The convex information bottleneck lagrangian.
\newblock \emph{Entropy}, 22\penalty0 (1):\penalty0 98, 2020.

\bibitem[Shamir et~al.(2010)Shamir, Sabato, and Tishby]{shamir2010learning}
Shamir, O., Sabato, S., and Tishby, N.
\newblock Learning and generalization with the information bottleneck.
\newblock \emph{Theoretical Computer Science}, 411\penalty0 (29-30):\penalty0 2696--2711, 2010.

\bibitem[Sicks et~al.(2021)Sicks, Korn, and Schwaar]{sicks2021generalised}
Sicks, R., Korn, R., and Schwaar, S.
\newblock A generalised linear model framework for $\beta$-variational autoencoders based on exponential dispersion families.
\newblock 2021.

\bibitem[Szegedy et~al.(2016)Szegedy, Vanhoucke, Ioffe, Shlens, and Wojna]{szegedy2016rethinking}
Szegedy, C., Vanhoucke, V., Ioffe, S., Shlens, J., and Wojna, Z.
\newblock Rethinking the inception architecture for computer vision.
\newblock In \emph{Proceedings of the IEEE conference on computer vision and pattern recognition}, pp.\  2818--2826, 2016.

\bibitem[Tishby \& Zaslavsky(2015)Tishby and Zaslavsky]{tishby2015deep}
Tishby, N. and Zaslavsky, N.
\newblock Deep learning and the information bottleneck principle.
\newblock In \emph{2015 ieee information theory workshop (itw)}, pp.\  1--5. IEEE, 2015.

\bibitem[Tishby et~al.(2000)Tishby, Pereira, and Bialek]{tishby2000information}
Tishby, N., Pereira, F.~C., and Bialek, W.
\newblock The information bottleneck method.
\newblock \emph{arXiv preprint physics/0004057}, 2000.

\bibitem[Tschannen et~al.(2018)Tschannen, Bachem, and Lucic]{tschannen2018recent}
Tschannen, M., Bachem, O., and Lucic, M.
\newblock Recent advances in autoencoder-based representation learning.
\newblock \emph{arXiv preprint arXiv:1812.05069}, 2018.

\bibitem[Vera et~al.(2018)Vera, Piantanida, and Vega]{vera2018role}
Vera, M., Piantanida, P., and Vega, L.~R.
\newblock The role of the information bottleneck in representation learning.
\newblock In \emph{2018 IEEE international symposium on information theory (ISIT)}, pp.\  1580--1584. IEEE, 2018.

\bibitem[Wu et~al.(2020)Wu, Fischer, Chuang, and Tegmark]{wu2020learnability}
Wu, T., Fischer, I., Chuang, I.~L., and Tegmark, M.
\newblock Learnability for the information bottleneck.
\newblock In \emph{Uncertainty in Artificial Intelligence}, pp.\  1050--1060. PMLR, 2020.

\bibitem[Xiao et~al.(2017)Xiao, Rasul, and Vollgraf]{xiao2017fashion}
Xiao, H., Rasul, K., and Vollgraf, R.
\newblock Fashion-mnist: a novel image dataset for benchmarking machine learning algorithms.
\newblock \emph{arXiv preprint arXiv:1708.07747}, 2017.

\bibitem[Yu et~al.(2021)Yu, Yu, and Pr{\'\i}ncipe]{yu2021deep}
Yu, X., Yu, S., and Pr{\'\i}ncipe, J.~C.
\newblock Deep deterministic information bottleneck with matrix-based entropy functional.
\newblock In \emph{ICASSP 2021-2021 IEEE International Conference on Acoustics, Speech and Signal Processing (ICASSP)}, pp.\  3160--3164. IEEE, 2021.

\bibitem[Zhang et~al.(2020)Zhang, Kailkhura, and Han]{zhang2020mix}
Zhang, J., Kailkhura, B., and Han, T. Y.-J.
\newblock Mix-n-match: Ensemble and compositional methods for uncertainty calibration in deep learning.
\newblock In \emph{International conference on machine learning}, pp.\  11117--11128. PMLR, 2020.

\end{thebibliography}
\bibliographystyle{icml2024}

\newpage
\appendix
\onecolumn
\section{Proofs}
\subsection{Proof of Lemma \ref{prop_optimal_point}}
\label{subsection_proof_lemma}
\begin{proof}
This Proof is based on the works for VAE \cite{lucas2019don, kumar2020implicit, sicks2021generalised}. See also these works. First, as a preparation, we perform some calculations related to the classifier $q_{\phi}$ in advance. For any $W \in \mathbb{R} ^{d \times \kappa}$, there exists $A \in \mathbb{R}^{d-1\times \kappa}$ that produces the same classifier $q_{\phi}$ when we set
\begin{equation}
W=
\begin{bmatrix}
A \\
 \boldsymbol{0}^{\top} \\
\end{bmatrix}
\end{equation}
Thus, we will seek the optimal solution for $A$ instead of $W$.
In this case, by defining $v(\boldsymbol z):= A \boldsymbol z$, the classifier $q_{\phi}$ can be represented as
\begin{equation}
\label{truncated_softmax}
q_{\phi}(\boldsymbol y|\boldsymbol z)=\frac{\exp(\Tilde{\boldsymbol y}^{\top}v(\boldsymbol z))}{1+\Sigma_{i=1}^{d-1}\exp(v_{i}(\boldsymbol z))}
\end{equation}
Here,  $v_{i}(\boldsymbol z)$ denotes the i-th component of the vector $v(\boldsymbol z)$. Additionally, $\Tilde{\boldsymbol y} \in \mathbb R^{d-1}$ follows Truncated multinomial distribution \cite{peyhardi2014new} and thus represents the first d-1 dimensions of  the one-hot vector $\boldsymbol y$. Therefore, the zero vector corresponds to the last category. 
Here, from Equation (\ref{truncated_softmax}), we can calculate as
\begin{equation}
\label{source3}
\log q_{\phi}(\boldsymbol y|\boldsymbol 0) = -\log d
\end{equation}
\begin{equation}
\label{source4}
\nabla_{\boldsymbol v}\log q_{\phi}(\boldsymbol y|\boldsymbol 0)= \Tilde{\boldsymbol y}-\frac{\boldsymbol1}{d}
\end{equation}
where $\boldsymbol 1$ denotes a vector in which all components are $1$. L.H.S of Equation (\ref{source4}) represents the gradient of $\log q_{\phi}(\boldsymbol y|\boldsymbol z)$ as a function of $\boldsymbol v := v(\boldsymbol{z})$ evaluated at $\boldsymbol{z}=\boldsymbol{0}$, which implies $\boldsymbol v= \boldsymbol{0}$. Using this, we derive the following expression for the gradient of $f_{\boldsymbol{y}}$.
\begin{equation}
\label{grad_f}
\nabla f_{\boldsymbol{y}}(\boldsymbol{0})=J_v(\boldsymbol{0})^{\top}\nabla_{\boldsymbol v} \log q_{\phi}(\boldsymbol y|\boldsymbol{0})= A^{\top}\rbra*{\Tilde{\boldsymbol{y}}-\frac{\boldsymbol1}{d}}
\end{equation}
where $J_v(\boldsymbol{0})$ is the Jacobian of $v(\boldsymbol{z})$.
And we define $\Gamma \in \mathbb{R} ^{(d-1) \times (d-1)}$ as
\begin{equation}
\label{gamma}
\Gamma:=-\nabla^2_{\boldsymbol v} \log q_{\phi}(\boldsymbol y|\boldsymbol{0})
\end{equation}
This expression represents the Hessian of the function $\log q_{\phi}(\boldsymbol y|\boldsymbol z)$ with respect to $\boldsymbol v$ when $\boldsymbol{z} = \boldsymbol{0}$. We can calculate $\Gamma$ as 
\begin{equation}
\Gamma=\frac{1}{d^2}
\begin{bmatrix}
d-1 &  & -1 \\
 & \ddots & \\
-1    &  & d-1 \\
\end{bmatrix}
\end{equation}
and its inverse matrix is  
\begin{equation}
\Gamma^{-1}=d
\begin{bmatrix}
2 &  & 1 \\
 & \ddots & \\
1    &  & 2 \\
\end{bmatrix}
\end{equation}
This matrix is positive definite, thus $\Gamma$ is also positive definite. The Hessian of $f_{\boldsymbol y}$ is represented as
\begin{equation}
\label{hessian}
\nabla^2f_y(\boldsymbol{0})=J_v(\boldsymbol{0})^{\top}(\nabla^2_{\boldsymbol v} \log q_{\phi}(\boldsymbol y|\boldsymbol 0))J_v(\boldsymbol{0})
=-A^{\top}\Gamma A
\end{equation}
Thus $\nabla^2f_y(\boldsymbol{0})$ is negative definite.

Next, using these result, we seek to find the optimal values for $\theta = \cbra{\mu, \Sigma} $ and $A$ when $\beta \in (0, 1]$. The case where $\beta=0$ is referred to later. We first fix $A$ and find optimal values for $\theta$.
The KL-Divergence in Equation (\ref{approx_objective}) under the settings here is represented as
\begin{equation}
\label{kl in objective}
D_{KL}\rbra*{p_{\theta}\rbra*{Z|x_i}||r(Z)}=\frac{1}{2}\rbra*{\text{tr}\rbra*{\Sigma\rbra*{x_i}}-\log \abs*{\Sigma\rbra*{x_i}} + \norm*{\mu  \rbra*{x_i}}_2^2-\kappa}
\end{equation}
From the expectation in Equation (\ref{approx_objective}), we have
\begin{equation}
\label{expectation in objective}
\begin{split}
\mathbb{E}_{p_\phi (Z|x_i)}\sbra*{T_{\boldsymbol y_i}(Z)}
&= \log q_{\phi}(\boldsymbol y_i|\boldsymbol 0)+ \mathbb{E}_{p_{\theta}(Z|x_i)}\sbra*{\nabla f_{\boldsymbol y_i}(\boldsymbol{0})^{\top}Z+\frac{1}{2}Z^{\top}\nabla^2f_{\boldsymbol y_i}(\boldsymbol{0})Z}\\
&= \log q_{\phi}(\boldsymbol y_i|\boldsymbol 0)+\nabla f_{\boldsymbol y_i}(\boldsymbol{0})^{\top}\mu(x_i) +\frac{1}{2}\text{tr}\rbra*{\nabla^2f_{\boldsymbol y_i}(\boldsymbol{0})\Sigma(x_i)}+\frac{1}{2}\text{tr}\rbra*{\nabla^2f_{\boldsymbol y_i}(\boldsymbol{0})\mu(x_i)\mu(x_i)^{\top}}
\end{split}
\end{equation}
From Equations (\ref{kl in objective}) and (\ref{expectation in objective}), we obtain
\begin{equation}
\label{approx_objective2}
\begin{split}
\hat{\mathcal{L}}_{VIB}(\theta, \theta; \beta) =& \frac{1}{N}\sum_{i=1}^N\left[-\frac{\beta}{2}\rbra*{\text{tr}\rbra*{\rbra*{I-\frac{1}{\beta}\nabla^2f_{\boldsymbol{y}_i}\rbra*{\boldsymbol{0}}}\Sigma\rbra*{x_i}}+\log \abs*{\Sigma\rbra*{x_i}^{-1}}}\right.\\
&\left.-\frac{\beta}{2}\norm*{\mu(x_i)}^2_2+\frac{\beta\kappa}{2}+\log q_{\phi}(\boldsymbol y_i|\boldsymbol{0})+\nabla f_{\boldsymbol{y}_i}(\boldsymbol{0})^{\top}\mu(x_i)+\frac{1}{2}\text{tr}\rbra*{\nabla^2 f_{\boldsymbol y_i}(\boldsymbol{0})\mu(x_i)\mu(x_i)^{\top}}\right]
\end{split}
\end{equation}
Here, we use the following proposition:
\vskip\baselineskip
\textit{If matrix $B$ is positive definite, then}
 \begin{equation}
\label{lemma2}
B = \underset{D \succ 0} {\operatorname{argmin}} \rbra*{\text{tr}\rbra*{BD^{-1}}+ \log \abs*{D}}
\end{equation}
\vskip\baselineskip
You can refer to Lemma 2 in \citet{sicks2021generalised} for the proof.
Using this and $I-\frac{1}{\beta}\nabla^2f_{\boldsymbol y_i}(\boldsymbol{0})\succ 0$ (remember that $\nabla^2f_y(\boldsymbol{0})$ is negative definite), by maximizing Equation (\ref{approx_objective2}) with respect to $\Sigma\rbra*{x_i}$, we obtain
\begin{equation}
\label{sigma_hat}
\hat{\Sigma}\rbra*{x_i} = \rbra*{I-\frac{1}{\beta}\nabla^2f_{\boldsymbol y_i}(\boldsymbol 0)}^{-1}
\end{equation}
Substituting this into Equation (\ref{approx_objective2}), we get
\begin{equation}
\label{mu_concave}
\begin{split}
\hat{\mathcal{L}}_{VIB}(\mu, \phi; \beta) =& \frac{1}{N}\sum_{i=1}^N\left[-\frac{\beta}{2}\log \abs*{\Sigma(x_i)^{-1}}-\frac{\beta}{2}\norm*{\mu(x_i)}^2_2+\log q_{\phi}(\boldsymbol y_i|\boldsymbol{0}) \right.\\
&\left.+\nabla f_{\boldsymbol{y}_i}(\boldsymbol{0})^{\top}\mu(x_i)+\frac{1}{2}\text{tr}\rbra*{\nabla^2 f_{\boldsymbol{y}_i}(\boldsymbol{0})\mu(x_i)\mu(x_i)^{\top}}\right]
\end{split}
\end{equation}
The stationary points with respect to $\mu(x_i)$ are given by
\begin{equation}
\label{mu_hat}
\hat{\mu}(x_i) = \frac{1}{\beta}\hat{\Sigma}(x_i)\nabla f_{\boldsymbol{y}_i}(\boldsymbol{0})
\end{equation}
Since Equation (\ref{mu_concave}) is concave with respect to $\mu(x_i)$,  this represents the global optimum.
Substituting this into Equation (\ref{mu_concave}), we get
\begin{equation}
\label{loss_for_theta}
\begin{split}
\hat{\mathcal{L}}_{VIB}(\phi; \beta) =& \frac{1}{N}\sum_{i=1}^N\left[\nabla f_{\boldsymbol{y}_i}(\boldsymbol{0})^{\top}E\nabla f_{\boldsymbol{y}_i}(\boldsymbol{0})-\frac{\beta}{2}\log \abs*{\Sigma(x_i)^{-1}}+\log q_{\phi}(\boldsymbol y_i|\boldsymbol{0}) \right]
\end{split}
\end{equation}
where 
\begin{equation}
E := -\frac{1}{2\beta}\hat{\Sigma}(x_i)^2+\frac{1}{2\beta^2}\hat{\Sigma}(x_i)\nabla^2 f_{\boldsymbol{y}_i}(\boldsymbol{0})\hat{\Sigma}(x_i) + \frac{1}{\beta}\hat{\Sigma}(x_i)
\end{equation}

Then, find the optimal values for $A$.
Using Equations (\ref{source3}) and (\ref{grad_f}), we can rewrite Equation (\ref{loss_for_theta}) as follows
\begin{equation}
\label{loss_for_theta_2}
\begin{split}
\hat{\mathcal{L}}_{VIB}(\phi; \beta) =& \frac{1}{N}\sum_{i=1}^N\left[\rbra*{\Tilde{\boldsymbol{y}}_i-\frac{\boldsymbol1}{d}}^{\top}\rbra*{AEA^{\top}-\frac{1}{2}\Gamma^{-1}}\rbra*{\Tilde{\boldsymbol{y}}_i-\frac{\boldsymbol1}{d}}\right.\\
&\left.+\frac{1}{2}\rbra*{\Tilde{\boldsymbol{y}}_i-\frac{\boldsymbol1}{d}}^{\top}\Gamma^{-1}\rbra*{\Tilde{\boldsymbol{y}}_i-\frac{\boldsymbol1}{d}}-\frac{\beta}{2}\log \abs*{\Sigma(x_i)^{-1}} - \log d \right]
\end{split}
\end{equation}
Since $\Gamma$ is positive definite, there exists $H \in \mathbb{R} ^{(d-1) \times (d-1)}$ such that $HH^{\top}=\Gamma$. Considering singular value decomposition, we have $H^{\top}A=U\Tilde{D}V^{\top}$, where $U \in \mathbb{R}^{(d-1) \times (d-1)}, V\in \mathbb{R}^{\kappa \times \kappa}$are orthogonal matrices. We will consider the case where $\kappa > d-1$, however please note that the same results for Equations (\ref{source1}) and (\ref{source2}) can also be obtained in the same manner when $\kappa \le d-1$. In this case, $\Tilde{D}$ is represented as
\begin{equation}
\Tilde{D}=
\begin{bmatrix}
\delta_1 &  & 0& \\
 & \ddots & &\text{\huge{0}}\\
0    &  & \delta_{d-1} &\\
\end{bmatrix}
\in \mathbb{R}^{(d-1) \times \kappa}
\end{equation}
Using Equation (\ref{hessian}), we have 
\begin{equation}
\begin{split}
\hat{\Sigma}(x_i)=\rbra*{V\rbra*{I+\frac{1}{\beta}\Tilde{D}^{\top}\Tilde{D}}V^{\top}}^{-1}=V\hat{D}V^{\top}
\end{split}
\end{equation}
where
\begin{equation}
\hat{D}:=\text{diag}\rbra*{\frac{1}{1+\beta^{-1}\delta_1^2},...,\frac{1}{1+\beta^{-1}\delta_{d-1}^2},1,...,1}\in \mathbb{R}^{\kappa \times \kappa}
\end{equation}
For $E$, we can calculate as
\begin{equation}
\begin{split}
E&=-\frac{1}{2\beta}\rbra*{\hat{\Sigma}(x_i)^2-\frac{1}{\beta}\hat{\Sigma}(x_i)\nabla^2 f_{\boldsymbol{y}_i}(\boldsymbol{0})\hat{\Sigma}(x_i) - 2\hat{\Sigma}(x_i)}\\
&=-\frac{1}{2\beta}V\rbra*{\hat{D}^2+\frac{1}{\beta}\hat{D}\Tilde{D}^{\top}\Tilde{D}\hat{D}-2\hat{D}}V^{\top}\\
&=\frac{1}{2\beta}V\hat{D}V^{\top}
\end{split}
\end{equation}
Then, we calculate $AEA^{\top}-\frac{1}{2}\Gamma^{-1}$ and $\log\abs*{\hat{\Sigma}(x_i)^{-1}}$ in Equation (\ref{loss_for_theta_2}) as follows.
\begin{equation}
\label{source1}
\begin{split}
AEA^{\top}-\frac{1}{2}\Gamma^{-1}&=\frac{1}{2\beta}AV\hat{D}V^{\top}A^{\top}-\frac{1}{2}\Gamma^{-1}\\
&=-\frac{1}{2}H^{-T}U\rbra*{-\frac{1}{\beta}\Tilde{D}\hat{D}\Tilde{D}^{\top}+I}U^{\top}H^{-1}\\
&=-\frac{1}{2}H^{-T}U\rbra*{-\frac{1}{\beta}\Tilde{D}\Tilde{D}^{\top}+I}^{-1}U^{\top}H^{-1}\\
&=-\frac{\beta}{2}\Gamma^{-1}\rbra*{\beta\Gamma+AA^{\top}}^{-1}\Gamma^{-1}
\end{split}
\end{equation}
\begin{equation}
\label{source2}
\begin{split}
\log\abs*{\hat{\Sigma}(x_i)^{-1}}&= \log\abs*{I+\frac{1}{\beta}\Tilde{D}^{\top}\Tilde{D}}\\
&= \log\abs*{I+\frac{1}{\beta}\Tilde{D}\Tilde{D}^{\top}}\\
&= \log\abs*{H^{-T}U\rbra*{I+\frac{1}{\beta}\Tilde{D}\Tilde{D}^{\top}}U^{\top}H^{-1}}+\log\abs*{\Gamma}\\
&= \log\abs*{\beta\Gamma^{-1}+AA^{\top}}+\log\abs*{\Gamma}-(d-1)\log\beta
\end{split}
\end{equation}
Substituting Equations (\ref{source1}) and (\ref{source2}) into Equation (\ref{loss_for_theta_2}), we have
\begin{equation}
\begin{split}
\hat{\mathcal{L}}_{VIB}(\phi; \beta) =&\frac{1}{N}\sum_{i=1}^N\sbra*{-\frac{\beta}{2}\rbra*{\Gamma^{-1}\rbra*{\Tilde{\boldsymbol{y}}_i-\frac{\boldsymbol1}{d}}}^{\top}C^{-1}\rbra*{\Gamma^{-1}\rbra*{\Tilde{\boldsymbol{y}}_i-\frac{\boldsymbol1}{d}}}+\frac{1}{2}\rbra*{\Tilde{\boldsymbol{y}}_i-\frac{\boldsymbol1}{d}}^{\top}\Gamma^{-1}\rbra*{\Tilde{\boldsymbol{y}}_i-\frac{\boldsymbol1}{d}}}\\
&-\frac{\beta}{2}\log|C|+\frac{\beta(d-1)}{2}\log \beta-\frac{\beta}{2}\log|\Gamma|-\log d
\end{split}
\end{equation}
where
\begin{equation}
C := \beta\Gamma^{-1}+AA^{\top}
\end{equation}
Furthermore, we obtain
\begin{equation}
\label{loss_for_theta3}
\hat{\mathcal{L}}_{VIB}(\phi; \beta) = -\frac{\beta}{2}\rbra*{\text{tr}(SC^{-1})+\log|C|}+a
\end{equation}
where $S$  is a constant matrix represented as
\begin{equation}
\label{S}
S := \frac{1}{N}\sum_{i=1}^N\rbra*{\Gamma^{-1}\rbra*{\Tilde{\boldsymbol{y}}_i-\frac{\boldsymbol1}{d}}}\rbra*{\Gamma^{-1}\rbra*{\Tilde{\boldsymbol{y}}_i-\frac{\boldsymbol1}{d}}}^{\top}
\end{equation}
and $a$ is also a constant 
\begin{equation}
a := \frac{\beta(d-1)}{2}\log \beta-\frac{\beta}{2}\log|\Gamma|-\log d +\frac{1}{2N}\sum_{i=1}^N\rbra*{\Tilde{\boldsymbol{y}}_i-\frac{\boldsymbol1}{d}}^{\top}\Gamma^{-1}\rbra*{\Tilde{\boldsymbol{y}}_i-\frac{\boldsymbol1}{d}}
\end{equation}
Under the assumption that class labels are balanced in the training data, Equation (\ref{S}) is calculated as
\begin{equation}
S=d
\begin{bmatrix}
2 &  & 1 \\
 & \ddots &\\
1    &  & 2 \\
\end{bmatrix}
=\Gamma^{-1}
\end{equation}
Using Equation (\ref{lemma2}) and the fact that $S$ and $C$ are positive definite, by maximizing Equation (\ref{loss_for_theta3}) with respect to $C$, we obtain
\begin{equation}
\hat{C}=S
\end{equation}
Thus, the optimal $A$ that realize this is represented as
\begin{equation}
\label{optimal_A}
\hat{A}\hat{A}^{\top}=(1-\beta)\Gamma^{-1}
\end{equation}
Satisfying this for any $\beta \in (0, 1]$ requires $\kappa \ge \text{rank} \, A = \text{rank} \, \Gamma^{-1} = d-1$, and conversely, when $\kappa \ge d-1$, there exists an $A$ that satisfies Equation  (\ref{optimal_A}). In the following, we will specifically consider the case where $\kappa=d-1$. Since $\Gamma^{-1} \succ 0$, there exists an $K \in \mathbb{R}^{(d-1)\times (d-1)}$ that satisfies
\begin{equation}
\label{K_def}
K^{\top}K:=\Gamma^{-1}
\end{equation}
This can be calculated by orthogonal diagonalization (see Appendix \ref{subsection_l} for details).
Thus, the optimal $W$ is represented as 
\begin{equation}
\label{w_hat}
\hat{W}=\sqrt{1-\beta}L^{\top}
\end{equation}
where
\begin{equation}
\label{Q_def}
L:=
\begin{bmatrix}
K \boldsymbol{0} \\
\end{bmatrix}
\end{equation}
Then, we calculate $\hat{\Sigma}, \hat{\mu}$ in this case. The hessian of $f_{\boldsymbol{y}_i}$is 
\begin{equation}
\label{opt_hessian}
\begin{split}
\nabla^2f_{\boldsymbol{y}_i}(\boldsymbol{0})=-\hat{A}^{-1}\hat{A}\hat{A}^{\top}\Gamma \hat{A}=-(1-\beta)I\\
\end{split}
\end{equation}
Substituting this  into Equation (\ref{sigma_hat}), we get
\begin{equation}
\label{sigma_hat2}
\hat{\Sigma}(x_i) = \beta I
\end{equation}
Substituting Equations (\ref{sigma_hat2}) and (\ref{grad_f}) into Equation (\ref{mu_hat}), we have
\begin{equation}
\label{mu_hat2}
\begin{split}
\hat{\mu}(x_i) = \nabla f_{\boldsymbol{y}_i}(\boldsymbol{0})=\sqrt{1-\beta}K\rbra*{\Tilde{\boldsymbol y}_i-\frac{\boldsymbol{1}}{d}}=\sqrt{1-\beta}L\rbra*{\boldsymbol y_i-\frac{\boldsymbol{1}}{d}}
\end{split}
\end{equation}
Up to this point, for $\beta \in (0, 1]$, the optimal values for $\mu, \Sigma$ and $W$ have been derived as described in Equations (\ref{mu_hat2}), (\ref{sigma_hat2})  and (\ref{w_hat}).

Finally, we consider the case where $\beta = 0$. From Equation (\ref{ce_taylor_approx}), we calculate $T_{\boldsymbol y_i}(Z)$ as follows.
\begin{equation}
\label{T_wrt_v}
T_{\boldsymbol y}(\boldsymbol{z})= v(\boldsymbol{z})^{\top}\rbra*{-\frac{1}{2}\Gamma}v(\boldsymbol{z})+\rbra*{\Tilde{\boldsymbol{y}}-\frac{\boldsymbol{1}}{d}}^{\top}v(\boldsymbol{z})-\log d
\end{equation}
This is maximized when $v(\boldsymbol{z})= \Gamma^{-1}\rbra*{\Tilde{\boldsymbol{y}}-\frac{\boldsymbol{1}}{d}}$. This condition is always satisfied when $\mu, \Sigma$ and $W$ correspond to Equations (\ref{mu_hat2}), (\ref{sigma_hat2})  and (\ref{w_hat}), respectively, with $\beta = 0$. Consequently, under these circumstances, the value of the expectation $\mathbb{E}_{p_\phi (Z|x_i)}\sbra*{T_{\boldsymbol y_i}(Z)}$ reaches its maximum.

In summary, for $\beta \in [0, 1]$, when $\mu, \Sigma$ and $W$ are aligned with Equations (\ref{mu_hat2}), (\ref{sigma_hat2})  and (\ref{w_hat}), respectively, the Taylor approximation of the VIB objective reaches its maximum.
\end{proof}

\subsection{Proof of Theorem \ref{prop_uniform_conv} and Theorem \ref{prop_mono_inc}}
\label{subsection_proof_thm1}
\begin{proof}
From Equation (\ref{approx_objective2}), we obtain
\begin{equation}
    \begin{split}
        & \hat{\mathcal{L}}_{VIB}\rbra*{\mu, \Sigma,W;\beta} = \frac{1}{N}\sum_{i=1}^N\sbra*{-\frac{\beta}{2}\norm*{\mu(x_i)}_2^2+\nabla f_{\boldsymbol{y}_i}(\boldsymbol{0})^{\top}\mu(x_i)+\frac{1}{2}\mu(x_i)^{\top}\nabla^2 f_{\boldsymbol y_i}(\boldsymbol{0})\mu(x_i)} +g_{\beta}(\Sigma,W)
    \end{split}
\end{equation}
where $g_{\beta}$ is a function of $\Sigma$ and $W$. When we use $ \Tilde{W}_{\beta}$, $\nabla f_{\boldsymbol{y}_i}(\boldsymbol{0})$ and $\nabla^2 f_{\boldsymbol y_i}(\boldsymbol{0})$ are calculated as Equations (\ref{mu_hat2}) and (\ref{opt_hessian}) respectively.
Setting $\mu$ as $\sqrt{1-\beta}h$ with some $h: \mathcal{X} \rightarrow \mathbb R^{d-1}$ and substituting these, we have
\begin{equation}
\label{func_of_h}
\begin{split}
        \hat{\mathcal{L}}_{VIB}\rbra*{\sqrt{1-\beta}h, \Tilde{\Sigma}_{\beta}, \Tilde{W}_{\beta};\beta} &= -\frac{1-\beta}{2N}\sum_{i=1}^N\sbra*{\norm*{h(x_i)}_2^2-2\rbra*{L\rbra*{\boldsymbol{y}_i-\frac{\boldsymbol{1}}{d}}}^{\top}h(x_i)} +g_{\beta}(\Tilde{\Sigma}_{\beta}, \Tilde{W}_{\beta})\\
        &=\frac{1-\beta}{2}\rbra*{-\frac{1}{N}\sum_{i=1}^N\norm*{h(x_i)-L\rbra*{\boldsymbol{y}_i-\frac{\boldsymbol{1}}{d}}}_2^2}+g_{\beta}(\Tilde{\Sigma}_{\beta}, \Tilde{W}_{\beta})+c_{\beta}
\end{split}
\end{equation}
where $c_{\beta}$ is a constant.
Thus, we have
\begin{equation}
\begin{split}
        \hat{\mathcal{L}}_{VIB}\rbra*{\Tilde{\mu}_{\beta, \psi}, \Tilde{\Sigma}_{\beta}, \Tilde{W}_{\beta};\beta} =\frac{1-\beta}{2}J_{FVIB}(\psi)+g_{\beta}(\Tilde{\Sigma}_{\beta}, \Tilde{W}_{\beta})+c_{\beta}
\end{split}
\end{equation}
This proves Theorem \ref{prop_mono_inc}.

Let $\hat{h}$ be a function satisfying $\hat{h}(x_i)=L\rbra*{\boldsymbol{y}_i-\frac{\boldsymbol{1}}{d}}$ for all $i = 1,2...,N$. From Lemma \ref{prop_optimal_point}, $\hat{\mathcal{L}}_{VIB}\rbra*{\sqrt{1-\beta}\hat{h}, \Tilde{\Sigma}_{\beta}, \Tilde{W}_{\beta};\beta}$ reaches its maximum. Therefore, we obtain
\begin{equation}
\label{uni_conv}
    \begin{split}
        & \sup_{\beta \in \sbra*{0, 1}}\abs*{\hat{\mathcal{L}}_{VIB}\rbra*{\Tilde{\mu}_{\beta, \psi_t}, \Tilde{\Sigma}_{\beta}, \Tilde{W}_{\beta};\beta}-\max_{\phi, \theta}\hat{\mathcal{L}}_{VIB}(\phi, \theta;\beta)}
        = \sup_{\beta \in \sbra*{0, 1}}\abs*{\frac{1-\beta}{2}J_{FVIB}(\psi_t)}
        = -\frac{1}{2}J_{FVIB}(\psi_t)
    \end{split}
\end{equation}
As $t\rightarrow \infty$, Equation (\ref{uni_conv}) converges to $0$, thereby proving Theorem \ref{prop_uniform_conv}.
\end{proof}

\subsection{Proof of Proposition \ref{prop_opt_conf} }
\label{subsection_proof_prop}
\begin{proof}
For the necessary knowledge, first refer to the initial paragraph of Appendix \ref{subsection_proof_lemma}.  From Equation (\ref{ce_taylor_approx}), we have
\begin{equation}
\label{T_wrt_v}
T_{\boldsymbol y}(\boldsymbol{z})= v(\boldsymbol{z})^{\top}\rbra*{-\frac{1}{2}\Gamma}v(\boldsymbol{z})+\rbra*{\Tilde{\boldsymbol{y}}-\frac{\boldsymbol{1}}{d}}^{\top}v(\boldsymbol{z})-\log d
\end{equation}
This is maximized only when $v(\boldsymbol{z})= \Gamma^{-1}\rbra*{\Tilde{\boldsymbol{y}}-\frac{\boldsymbol{1}}{d}}$. By computing this and substituting into Equation (\ref{truncated_softmax}), we find that the corresponding class prediction matches the class of $\boldsymbol{y}$ and the confidence is $\frac{\exp(d)}{\exp(d)+d-1}$.
\end{proof}

\section{Implementation and Experimental Details}
\subsection{How to calculate $L$}
\label{subsection_l}
From Equation (\ref{Q_def}),
\begin{equation}
L:=
\begin{bmatrix}
K \boldsymbol{0} \\
\end{bmatrix}
\end{equation}
where
\begin{equation}
K^{\top}K := \Gamma^{-1} = d
\begin{bmatrix}
2 &  & 1 \\
 & \ddots &\\
1    &  & 2 \\
\end{bmatrix}
\in \mathbb R^{(d-1)\times (d-1)}
\end{equation}
By orthogonal diagonalization, we have $\Gamma^{-1}=:V D V^{\top}$. Since $\Gamma^{-1}$ is positive definite, we can consider a diagonal matrix whose diagonal elements are the square roots of those of D, denoted as $\sqrt{D}$. Then, we calculate $K$ as $K = (V \sqrt{D})^{\top}$. In the experiments, we utilize the NumPy package to calculate the eigenvalues and eigenvectors required for the orthogonal diagonalization.

\subsection{Experimental Setup}
\label{subsection_setup}
All experiments are conducted using PyTorch. The experimental setup is detailed in Table \ref{tab:encoder}. All models use a classifier composed of a single dense layer on top of the feature extractor. For MNIST and Fashion-MNIST,  the architecture and learning settings are based on \cite{alemi2016deep}. The architecture for CIFAR-10 and SVHN is adapted from \cite{achille2018information}. For LTAF, the architecture and learning settings are based on \cite{faust2018automated}.


\begin{table*}[h]
\caption{Experimental Setup. In the description of the feature extractor, the subscripts of each module indicate the output dimension, kernel size, and stride, in that order.}
\label{tab:encoder}
\vskip 0.15in
\begin{center}
\begin{small}
\begin{sc}
\begin{tabular}{l|lll}
\toprule
&MNIST, Fashion-MNIST&CIFAR-10, SVHN&LTAF \\
\midrule
&$x \in \mathbb{R}^{28 \times 28}$& $x \in \mathbb{R}^{3 \times 32 \times 32}$ &$x \in \mathbb{R}^{1\times100}$\\
&$ \rightarrow flatten \rightarrow Linear_{1024}$  & $ \rightarrow Conv_{96, 3, 1}$&$\rightarrow bidirectional\: LSTM_{400}$\\
&$\rightarrow ReLU \rightarrow Linear_{1024} $     &$\rightarrow ReLU \rightarrow Conv_{96, 3, 1}$&$\rightarrow temporal\: max$  \\
&$\rightarrow ReLU \rightarrow Linear_{\kappa\:or\:2\kappa}$    &$\rightarrow ReLU \rightarrow Conv_{96, 3, 2}$&$\rightarrow ReLU \rightarrow Linear_{50}$ \\
&                                                  &$\rightarrow ReLU \rightarrow Conv_{192, 3, 1}$&$\rightarrow ReLU \rightarrow Linear_{\kappa\:or\:2\kappa}$\\
Feature extractor&                                                  &$\rightarrow ReLU \rightarrow Conv_{192, 3, 1}$\\
&                                                  &$\rightarrow ReLU \rightarrow Conv_{192, 3, 2}$\\
&                                                  &$\rightarrow ReLU \rightarrow Conv_{192, 3, 1}$\\
&                                                  &$\rightarrow ReLU \rightarrow Conv_{192, 1, 1}$\\
&                                                  &$\rightarrow ReLU \rightarrow Conv_{\kappa \: or \: 2 \kappa, 1, 1}$\\
&                                                  &$\rightarrow spacial \: average$\\
\midrule
$\kappa$ in non-FVIB models&256&256&50\\
Epochs&200&200&50\\ 
Optimizer & Adam \cite{kingma2014adam} &Adam &Adam\\
Initial learning rate &$1.0 \times10^{-4}$&$1.0 \times 10^{-3}$&$1.0 \times 10^{-3}$\\
\multirow{2}{*}{Learning rate schedule}&Multiplied by 0.97 &Multiplied by 0.5&\multirow{2}{*}{Not used}\\
&every 2 epochs&at epochs 80, 120, and 160&\\

\bottomrule
\end{tabular}
\end{sc}
\end{small}
\end{center}
\vskip -0.1in
\end{table*}


\end{document}